\documentclass[10pt,twocolumn,letterpaper]{article}

\usepackage{cvpr}
\usepackage{times,paralist}
\usepackage{epsfig}
\usepackage{graphicx}
\usepackage{amsmath}
\usepackage{amssymb}

\usepackage{algorithm,algorithmic}
\usepackage{subcaption}

\usepackage[title]{appendix}

\newcommand{\R}{\mathbb{R}}
\newcommand{\X}{\mathbf{X}}
\newcommand{\x}{\mathbf{x}}
\newcommand{\Y}{\mathbf{Y}}
\newcommand{\y}{\mathbf{y}}
\newcommand{\Z}{\mathbf{Z}}
\newcommand{\z}{\mathbf{z}}
\newcommand{\h}{\mathbf{h}}
\newcommand{\uu}{\mathbf{u}}
\newcommand{\vv}{\mathbf{v}}
\newcommand{\qq}{\mathbf{q}}
\newcommand{\rr}{\mathbf{r}}
\newcommand{\s}{\mathbf{s}}
\def\ourmodel{CRow}

\usepackage[pagebackref=true,breaklinks=true,letterpaper=true,colorlinks,bookmarks=false]{hyperref}

 \cvprfinalcopy 

\def\cvprPaperID{2935} 

\begin{document}

\title{Conditional Recurrent Flow: Conditional Generation of Longitudinal Samples with Applications to Neuroimaging}

\author{\hspace{10pt}Seong Jae Hwang\\
	\hspace{10pt}Dept. of Computer Sciences\\
	\hspace{10pt}University of Wisconsin-Madison\\
	\hspace{10pt}{\tt\small sjh@cs.wisc.edu}
	\and \hspace{30pt}Zirui Tao\\
	\hspace{30pt}Dept. of Computer Sciences\\
	\hspace{30pt}University of Wisconsin-Madison\\
	\hspace{30pt}{\tt\small ztao23@wisc.edu}
	\and
	Won Hwa Kim\\
	Dept. of Computer Science and Engineering\\
	University of Texas, Arlington\\
	{\tt\small won.kim@uta.edu}
	\and
	Vikas Singh\\
	Dept. of Computer Sciences\\
	Dept. of Biostat. \& Med. Informatics\\
	University of Wisconsin-Madison\\
	{\tt\small vsingh@biostat.wisc.edu}
}


\maketitle
\begin{abstract}
	\vspace{-10pt}
	
	Generative models using neural network have opened a door to large-scale studies for various application domains,  
	especially for studies that suffer from lack of real samples to obtain statistically robust inference. 
	Typically, these generative models would train on existing data to 
	learn the underlying distribution of the measurements (e.g., images) in latent spaces conditioned on covariates (e.g., image labels), 
	and generate independent samples that are identically distributed in the latent space. 
	Such models may work for cross-sectional studies, however, 
	they are not suitable to generate data for {\em longitudinal} studies 
	that focus on ``progressive'' behavior in a sequence of data.  
	In practice, this is a quite common case in various neuroimaging studies whose goal is to characterize a trajectory of pathologies of a specific disease even from early stages.  
	This may be too ambitious especially when the sample size is small (e.g., up to a few hundreds). 
	Motivated from the setup above, we seek to develop a conditional generative model for longitudinal data generation 
	by designing an invertable neural network. 
	Inspired by recurrent nature of longitudinal data, we propose a novel neural network that incorporates recurrent subnetwork and context gating 
	to include smooth transition in a sequence of generated data.  
	Our model is validated on a video sequence dataset and a longitudinal AD dataset with various experimental settings 
	for qualitative and quantitative evaluations of the generated samples.  
	The results with the AD dataset captures AD specific group differences with sufficiently generated longitudinal samples that are consistent with existing literature, 
	which implies a great potential to be applicable to other disease studies.  

\end{abstract}
\section{Introduction}

Consider a dataset of {\em longitudinal or temporal sequences} of data samples $\{\x^t\}_{i=1}^N$ where each sample $\x_i$ comes with \textit{sequential covariates} $\{\y^t\}_{i=1}^N$, one
for each time point $t$. 
In other words, we assume that for each sequential sample $i$, $\x^1_i,\cdots,\x^T_i=\{\x^t\}_i$, 
the sequential covariates $\y^1_i,\cdots,\y^T_i=\{\y^t\}_i$ provide some pertinent auxiliary information associated with that sequential sample.
For example, 
in a neuroimaging study, if the sequential samples correspond to several longitudinal image scans of a participant over multiple years, the sequential covariate associated with each time point may be an assessment of disease severity or some other clinical measurement.
If the sequential samples denote written musical notes of a rhythm (e.g., tabla or drum beats), the sequential labels may specify the speed or frequency of the beats in a musical arrangement. 
If the sequential data corresponds to heart rate sensors when a participant is 
watching a video, the sequential covariate may indicate the presence 
of violence in the corresponding video segment. 

Our high level goal is to design conditional generative models for such sequential data. In particular, we want a model 
which provides us a type of flexibility that is highly desirable in this setting. 
For instance, for a sample drawn from 
the distribution after the generative model has been estimated, 
we should be able to ``adjust" the sequential covariates, say at a time point $t$, dynamically to influence the expected future predictions after $t$ for {\em that} sample. It makes sense that 
for a heart rate sequence, the appropriate sub-sequence should be 
influenced by {\em when} the ``violence" stimulus was 
introduced {\em as well as} the default heart rate pattern of the specific 
sample (participant) \cite{akselrod1981power}. 
Notice that 
when $T=1$, this construction is similar to conditional generative models where the ``covariate" or condition $\y$ may simply denote an attribute that we may want to adjust for a sample: for example, increase the smile or age attribute for a face image sampled from the distribution as in \cite{kingma2018glow}. 

We want our formulation to provide a modified set of $\x^t$s adaptively, if we adjust one or more sequential covariates $\y^t$s for that sample. 
If we know some important clinical information at some point during the study (say, at $t=5$), this information should influence the future generation $\x^{t>5}$ conditioned both on this sequential covariate or event $\y^5$ as well as the past sequence of this sample $\x^{t<5}$.
This will require \textit{conditioning} on the corresponding sequential covariates at \textit{each} time point $t$ by accurately capturing the posterior distribution $p(\x^t | \y^t)$.
This type of \textit{conditional sequential generation}, in effect, requires a generative model for sequential data which can dynamically incorporate time-specific sequential covariates $\y^t$ of interest to adaptively modify sequences. 

The above 
setup models a number of applications in 
in medical imaging, computer vision and engineering,  
that may need generation of frame sequences conditioned on frame-level covariates. 
In neuroimaging, many longitudinal studies focus on identifying disease trajectories \cite{alexander2002longitudinal,baddeley1991decline,landin2017disease}: for example, at what point in the future will the brain or specific regions in the brain exceed a threshold for brain atrophy? 
The future trend is invariably a function of 
clinical measurements that a participant provides at each visit {\em as well as} the past trend of the subject. 
From a methodological standpoint, constructing a sequential generative model may appear feasible by appropriately augmenting the generation process using existing generative models.
For example, it seems that 
one could simply concatenate the sequential measurements $\{\x^t\}$ as a single input for existing non-sequential conditional generative models
such as conditional GANs \cite{mirza2014conditional,isola2017image} and conditional variational autoencoders \cite{sohn2015learning,abbasnejad2017infinite}.


We find that for our application, an attractive alternative to  discriminator-generator based GANs, 
is a family of neural networks called normalizing flow \cite{rippel2013high,rezende2015variational,dinh2016density,dinh2014nice} 
which involve \textit{invertible networks} (i.e., reconstruct input from its output). What is particularly 
relevant is that such formulations work well for 
{\em conditionally} generating diverse samples with controllable degrees of freedom \cite{ardizzone2018analyzing} -- with an explicit mechanism to adjust the conditioning 
variable (or covariate). 
But the reader will notice that while these models, in principle, can be used to approximate the posterior probability given an input of any dimensions, 
concatenating a series of sequential 
inputs quickly blows up the size for these highly expressive models and quickly renders them impractical to run, even on high end GPU compute clusters.
Even if we optimistically assume computational feasibility, variable length sequences cannot easily be adapted to these innately non-sequential generative models, especially for those that extend beyond the training sequence length. 
Also, data generated in this manner involve simply ``concatenated" sequential data and do not take into account the innate temporal relationships among the sequences, fundamental in the success of recurrent models. 
These are the core issues we study here. 

Given various potential downstream applications and the issues identified above with conditional sequential generation problem, 
we seek a model which  
(i) efficiently generates high dimensional sequence samples of variable lengths 
(ii) with dynamic time-specific conditions reflecting upstream observations 
(iii) with fast posterior probability estimation.
We tackle the foregoing issues by introducing an invertible recurrent neural network, {\bf CRow}, that includes {\bf recurrent sub-network} and {\bf temporal context gating}. These 
modifications are critical in the following sense. 
{\em Invertibility} lets us precisely estimate the distribution of $p(\x^t | \y^t)$ in a latent space. 
Introducing {\em recurrent subnetworks} and {\em temporal context gating} enables 
obtaining cues from previous time points $\x^{<t}$ to generate temporally sensible subsequent time points $\x^{\ge t}$.
%
Specifically, our {\bf contributions} are: \begin{inparaenum}[\bfseries (A)]
\item Our model generates conditional sequential samples $\{\x^t\}$ given sequential covariates $\{\y^t\}$ for $t=1,\dots,T$ time points where $T$ can be arbitrarily long. Specifically, we allow this by posing the task as a \textit{conditional sequence inverse problem} based on a conditional invertible neural network \cite{ardizzone2018analyzing}.
\item Assessing the quality of the generated samples may not be trivial for certain modalities (e.g., non-visual features). With the specialized capability of the normalizing flow construction, our model estimates the posterior probabilities $p(\x^t | \y^t)$ of the generated sequences at each time point for potential downstream analyses involving uncertainty. 
\item We demonstrate an interesting practical application of our model in a longitudinal neuroimaging dataset. In an intuitive 
manner, we show that the generated longitudinal brain pathology trajectories can lead to identifying specific regions in the brain which are statistically associated with the manifestation of Alzheimer's disease.
\end{inparaenum}
\section{Preliminary: Invertible Neural Networks}
We first describe an \textit{invertible neural network} (INN) which inverts an output back to its input for solving inverse problems (i.e., $\z = f(\x) \Leftrightarrow \x = f^{-1}(\z)$).
This becomes the building block of our method; thus, before we present our main model, let us briefly describe a specific type of invertible structure which was originally specialized for density estimation with neural network models.

\subsection{Normalizing Flow}
Estimating the density $p_\X(\x)$ of sample $\x$ is a classical statistical problem in various fields including computer vision and machine learning in, e.g., uncertainty estimation \cite{gal2015bayesian,gal2016dropout}.
For tractable computation throughout the network, Bayesian adaptations are popular \cite{ranganath2015deep,fortunato2017bayesian,papamakarios2016fast,kingma2015variational,kendall2017uncertainties}, but these methods make assumptions on the prior distributions (e.g., exponential families).

A \textit{normalizing flow} \cite{rippel2013high,rezende2015variational}, first learns a function $f(\cdot)$ which maps a sample $\x$ to $\z = f(\x)$ where is from a standard normal distribution $\Z$. 
Then, with a change of variables formula, we estimate 
\begin{equation*}
\small
p_\X(\x) = p_\Z(\z) / |J_\X|, \quad J_\X = \left| \frac{\partial [\x = f^{-1}(\z)]}{\partial \z} \right|
\end{equation*}
where $J_\X$ is a Jacobian determinant.
Thus, $f(\cdot)$ must be invertible, i.e., $\x = f^{-1}(\z)$, and to use a neural network as $f(\cdot)$, a coupling layer structure was introduced in Real-NVP \cite{dinh2014nice,dinh2016density} for an easy inversion and efficient $J_\X$ computation as follows.

\begin{figure}[!t]
    \centering
    \begin{subfigure}[t]{0.49\columnwidth}
        \centering
        \includegraphics[width=0.95\columnwidth]{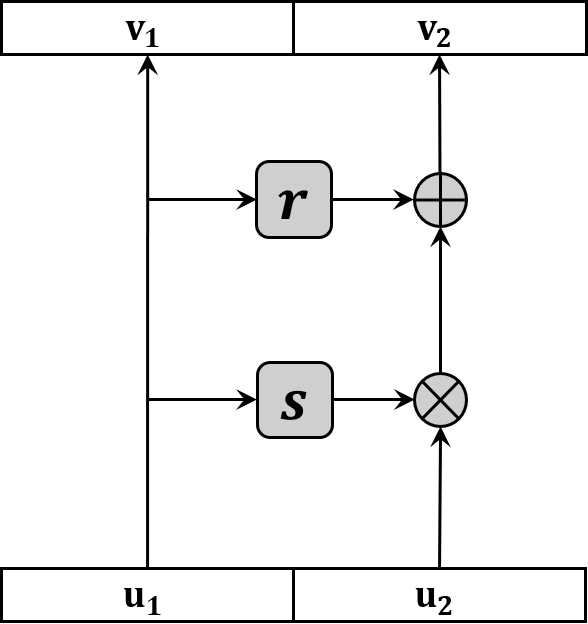}
        \caption{\footnotesize Forward map (Eq.~\eqref{eq:real-nvp})}
        \label{fig:coupling-layer-forward}
    \end{subfigure}%
    ~ 
    \begin{subfigure}[t]{0.49\columnwidth}
        \centering
        \includegraphics[width=0.95\columnwidth]{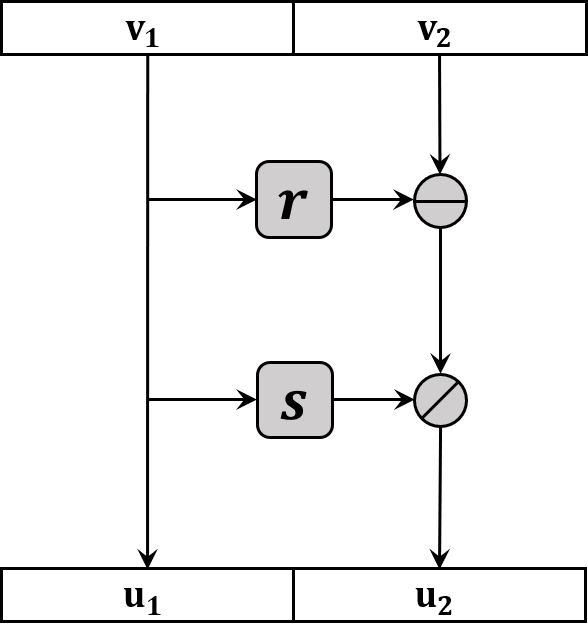}
        \caption{\footnotesize Inverse map (Eq.~\eqref{eq:inverse-real-nvp})}
        \label{fig:coupling-layer-inverse}
    \end{subfigure}
    \caption{Coupling layer in normalizing flow. Note the change of operation orders: $\uu \rightarrow \vv$ in forward and $\vv \rightarrow \uu$ in inverse.}
    \label{fig:coupling-layer}
\end{figure}

Without loss of generality, in the context of network structures, we use an input $\uu \in \R^d$ and an output $\vv \in \R^d$. First, we split $\uu$ into $\uu_1 \in \R^{d_1}$ and $\uu_2 \in \R^{d_2}$ where $d = d_1 + d_2$. Then, we forward map $\uu_1$ and $\uu_2$ to $\vv_1$ and $\vv_2$ respectively (Fig.~\ref{fig:coupling-layer-forward}):
\begin{equation}\label{eq:real-nvp}
\small
    \vv_1 = \uu_1, \quad \vv_2 = \uu_2 \otimes \exp (s(\vv_1)) + r(\vv_1)
\end{equation}
where $s, r$ are independent functions (i.e., subnetworks), and $\otimes$ and $+$ are element-wise product and addition respectively.
A straightforward arithmetic allows an exact inverse from $\vv$ to $\uu$ (Fig.~\ref{fig:coupling-layer-inverse}):
\begin{equation}\label{eq:inverse-real-nvp}
\small
    \uu_1 = \vv_1, \quad \uu_2 = (\vv_2 - r(\vv_1)) \oslash \exp(-s(\vv_1)) 
\end{equation}
where the subnetworks are \textit{identical} to those used in the forward process in Eq.~\eqref{eq:real-nvp}, and $\oslash$ and $-$ are element-wise division and subtraction respectively.
Not that the subnetworks are \textit{not} explicitly inverted, thus any arbitrarily complex network can be utilized.
Also, the Jacobian matrix $\partial \vv / \partial \uu$ is triangular so its determinant $J_\vv$ is just the product of diagonal entries (i.e., $\prod_i \exp(-s(\vv_1))_i)$ which is extremely easy to compute
(we will discuss this further in Sec.~\ref{sec:tgc}).

Then, to apply a transform on the ``bypassed'' $\uu_1$ as well, a \textit{coupling block} (consisting of two complementary coupling layers) is usually constructed:
\begin{equation}
\small
\begin{aligned}\label{eq:coupling}
    \vv_1 &= \uu_1 \otimes \exp (s_2(\uu_2)) + r_2(\uu_2) \\
    \vv_2 &= \uu_2 \otimes \exp (s_1(\vv_1)) + r_1(\vv_1)
\end{aligned}
\end{equation}
and its inverse
\begin{equation}
\small
\begin{aligned}\label{eq:inverse-coupling}
    \uu_2 &= (\vv_2 - r_1(\vv_1)) \oslash \exp(-s_1(\vv_1)) \\
    \uu_1 &= (\vv_1 - r_2(\uu_2)) \oslash \exp (-s_2(\uu_2)).
\end{aligned}
\end{equation}
Such a series of transformations allow a more complex mapping with a chain of efficient Jacobian determinant computations, i.e., $\det(AB) = \det(A)\det(B)$. More details are included in the appendix.

\section{Methods}
In this section, we describe our conditional sequence generation method called Conditional Recurrent Flow (\textit{\ourmodel{}}). 
We first describe a conditional invertible neural network (cINN) \cite{ardizzone2018analyzing} which is one component of our model. Then, we explain how to incorporate temporal context gating and discuss the settings where \ourmodel{} can be useful.

\subsection{Conditional Sample Generation}
Naturally, an inverse problem can be posed as a sample generation procedure by sampling a latent variable $\z$ and inverse mapping it to $\x = f^{-1}(\z)$.
The most critical concern is that we cannot specifically `choose' to generate an $\x$ of interest since a latent variable $\z$ does not provide any interpretable associations with $\x$.

In other words, estimating the \textit{conditional probability} $p(\x | \y)$ is desirable since it represents an underlying phenomenon of the input $\x \in \R^d$ and covariate $\y \in \R^k$ (e.g., the probability of a specific brain imaging measure $\x$ of interest given a diagnosis $\y$).
In fact, when we cast this problem into a normalizing flow, the goal becomes to construct an invertible network $f$ which maps a given input $\x \in \R^d$ to its corresponding covariate/label $\y \in \R^l$ and its latent variable $\z \in \R^m$ such that $[\y,\z] = f(\x)$, and the mapping must have an inverse $\x = f^{-1}([\y,\z])$ to be recovered.


Specifically, when a flow-based model jointly encodes latent and label information 
(i.e., $[\y,\z] = \vv = f(\x)$ via Eq.~\eqref{eq:coupling})
while ensuring that $p(\y)$ and $p(\z)$ are independent,
then the network becomes conditionally invertible.
Such network can be theoretically constructed through a bidirectional-type training \cite{ardizzone2018analyzing}, and this
allows a conditional sampling $\x = f^{-1}([\y,\z])$ and the posterior estimation $p(\x | \y)$.
This training process involves several losses: (1) $\mathcal{L}_\Z(p(\y,\z), p(\y)p(\z))$ enforces the independence between $p(\y,\z)$ and $p(\y)p(\z)$. (2) $\mathcal{L}_\Y(\y, \hat{\y})$ is the supervised label loss between our prediction $\hat{\y}$ and the ground truth $\y$. (3) $\mathcal{L}_\X(p(\x),p_\X)$ improves the likelihood of the input $\x$ with respect to the prior $p_\X$. $\mathcal{L}_\Z$ and $\mathcal{L}_\X$ are based on a kernel-based moment matching measure called Maximum Mean Discrepancy (MMD) \cite{dziugaite2015training}. See supplement for more details.

In practice, $\x$ and $[\y,\z]$ may not be of the same dimensions. To construct a square triangular Jacobian matrix, zero-padding both $\x$ and $[\y,\z]$ can alleviate this issue while also increasing the intermediate subnetwork dimensions for higher expressive power.
Also, the forward mapping is essentially a prediction task that we encounter often in computer vision and machine learning, i.e., predicting $\y = f(\x)$ or maximizing the likelihood $p(\y | \x)$ without explicitly utilizing the latent $\z$.
On the other hand, the inverse process of deriving $\x = f^{-1}(\y)$, allows a more scientifically based analysis of the underlying phenomena, e.g., the interaction between brain and observed cognitive function.

\begin{figure}[!t]
    \centering
    \includegraphics[width=0.9\columnwidth]{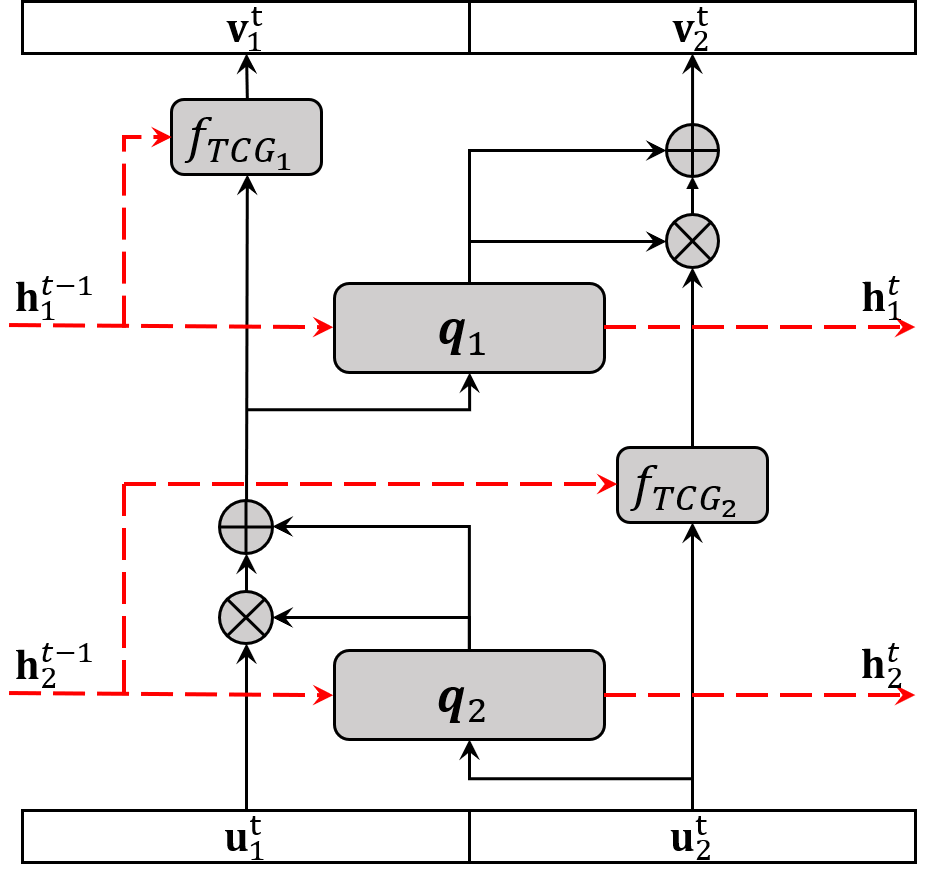}
    \caption{Our conditional sequence generation model. Only the forward map of a single block (two coupling layers) is shown for brevity. The inverse map involves a similar order of operations (analogous to Fig.~\ref{fig:coupling-layer-forward} and Fig.~\ref{fig:coupling-layer-inverse})}
    \label{fig:network}
\end{figure}

\subsection{Conditional Recurrent Flow}
The existing normalizing flow type networks cannot explicitly incorporate sequential data which are now increasingly becoming important in various applications.
Successful recurrent models such as gated recurrent unit (GRU) \cite{chung2014empirical,tang2015document} and Long short-term memory (LSTM) \cite{hochreiter1997long,sak2014long} explicitly focus on encoding the ``memory'' about the past time points and output proper state information for accurate sequential predictions given the past.
Similarly, generated sample sequences must also follow sequentially sensible patterns or trajectories resembling likely sequences by encoding appropriate temporal information for the subsequent time points. 

In order to overcome the issues above, we introduce Conditional Recurrent flow (\ourmodel{}) model for conditional sequence generation. Given a sequence of input/output pairs $\{\uu^t,\vv^t\}$ for $t=1,\dots,T$ time points, modeling the relationship between the variables across time needs to also account for the temporal characteristic of the sequence.
Variants of recurrent neural networks (RNN) such as GRU and LSTM have been showing success in numerous sequential problems, but they only enable forward mapping $\uu \rightarrow \vv$.
We are specifically interested in constructing an invertible network which is also recurrent such that given a sequence of inputs $\{\uu^t\}$ (i.e., $\{\x^t\}$) and their outputs $\{\vv^t\}$ (i.e., $\{\y^t\}$), we can successfully model the invertible relationship between those sequences.
We now show our conditional recurrent flow model called \textit{\ourmodel{}} which allows the conditional mapping between the \textit{sequence} of input $\{\uu^t\}$ and output $\{\vv^t\}$ where $\{\vv^t\} = \{\y^t,\z^t\}$ for posterior estimation and conditional sequence generation.

Without loss of generality, we can describe our model in terms of generic $\{\uu^t\}$ and $\{\vv^t\}$.
We follow the coupling block described in Eq.~\eqref{eq:coupling} and Eq.~\eqref{eq:inverse-coupling} to setup a normalizing flow type invertible model.
Then, we impose the recurrent nature to the model by allowing the model to learn and pass a hidden state $\h^t$ to the next time point through the recurrent subnetworks.
Specifically, we construct a \textit{recurrent subnetwork} $q$ which contains a recurrent network (i.e., GRU) internally.
This allows $q$ to take the previous hidden state $\h^{t-1}$ and output the next hidden state $\h^t$ as $[\qq, \h^t] = q(\uu, \h^{t-1})$ where $\qq$ is an element-wise transformation vector derived from $\uu$ analogous to the output of $s(\uu)$.
In previous coupling layers, two transformation vectors $\s = s(\cdot)$ and $\rr = r(\cdot)$  were explicitly computed from two subnetworks for each layer.
For \ourmodel{}, we follow the structure of Glow \cite{kingma2018glow} which computes a single vector $\qq = q(\cdot)$ and split it as $[\s, \rr] = \qq$.
This allows us to use a single hidden state while concurrently learning $[\s, \rr]$ which we denote as $\s = q_s(\cdot)$ and $\rr = q_r(\cdot)$ to indicate the corresponding vectors.
Thus, at each $t$ with given $[\uu^t_1, \uu^t_2] = \uu^t$ and $[\vv^t_1, \vv^t_2] = \vv^t$ our model becomes as follows:
\begin{equation}
\small
\begin{aligned}\label{eq:rec-coupling}
    \vv^t_1 &= \uu^t_1 \otimes \exp (q_{s_2}(\uu^t_2, \h^{t-1}_2)) + q_{r_2}(\uu^t_2, \h^{t-1}_2) \\
    \vv^t_2 &= \uu^t_2 \otimes \exp (q_{s_1}(\vv^t_1, \h^{t-1}_1)) + q_{r_1}(\vv^t_1, \h^{t-1}_1)
\end{aligned}
\end{equation}
and the inverse is
\begin{equation}
\small
  \begin{aligned}\label{eq:inverse-rec-coupling}
    \uu^t_2 &= (\vv^t_2 - q_{r_1}(\vv^t_1, \h^{t-1}_1)) \oslash \exp (q_{s_1}(\vv^t_1, \h^{t-1}_1)) \\
    \uu^t_1 &= (\vv^t_1 - q_{r_2}(\uu^t_2, \h^{t-1}_2)) \oslash \exp (q_{s_2}(\uu^t_2, \h^{t-1}_2)).
\end{aligned}  
\end{equation}
Note that the hidden states $\h^t_1$ and $\h^t_2$ generated from the RNN components of the subnetworks are implicitly used within the subnetwork architecture (i.e., inputs to additional fully connected layers) and also passed down to their corresponding RNN component in the next time point as shown in Fig.~\ref{fig:network}.

\subsubsection{Temporal Context Gating}\label{sec:tgc}
A standard (single) coupling layer transforms only a part of the input (i.e., $\uu_1$ in Eq.~\eqref{eq:real-nvp}) by design for the following triangular Jacobian matrix:
\begin{equation}\label{eq:jacobian}
\small
    J_\vv = \frac{\partial \vv}{\partial \uu} = \begin{vmatrix} \frac{\partial \vv_1}{\partial \uu_1} & \frac{\partial \vv_1}{\partial \uu_2} \\
    \frac{\partial \vv_2}{\partial \uu_1} & \frac{\partial \vv_2}{\partial \uu_2}
    \end{vmatrix}
    = \begin{vmatrix}
    I & 0 \\
    \frac{\partial \vv_2}{\partial \uu_1} & diag(\exp s(\uu_1))
    \end{vmatrix}
\end{equation}
thus $J_\vv = \exp(\sum_i (s(\uu_1))_i)$.
This is a result from Eq.~\eqref{eq:coupling} the element-wise operations on $\uu_2$ for diagonal ${\partial \vv_2}/{\partial \uu_2}$  and (2) the bypassing of $\uu_1$ for ${\partial \vv_1}/{\partial \uu_1} = I$ and ${\partial \vv_1}/{\partial \uu_2}  = 0$.
Ideally, transforming $\uu_1$ would be beneficial as well but avoided by the coupling layer design since an effective transformation cannot be learned from neither $\uu_1$ nor $\uu_2$ directly.
In case of \ourmodel{}, it incorporates a hidden state $\h^{t-1}$ from the previous time point which is \textit{not} a part of the variables which we model the relationship of (i.e., $\uu$ and $\vv$).
This is our recurrent information which adjust the mapping function $f(\cdot)$ to allow more accurate mapping depending on the previous sequences which is crucial for sequential modeling.

Specifically, we incorporate a \textit{temporal context gating} $f_{\textrm{TCG}}(\alpha^t, \h^{t-1})$ using the temporal information $\h^{t-1}$ on a given input $\alpha^t$ at $t$ as follows:
\begin{equation}
\small
\begin{aligned}\label{eq:cgate}
    f_{\textrm{TCG}}(\alpha^t, \h^{t-1}) &= \alpha^t \otimes cgate(\h^{t-1})   \quad \textrm{(forward)}\\
    f^{-1}_{\textrm{TCG}}(\alpha^t, \h^{t-1}) &= \alpha^t \oslash cgate(\h^{t-1})   \quad \textrm{(inverse)}
\end{aligned}
\end{equation}
where $cgate(\h^{t-1})$ can be any learnable function with a sigmoid function at the end.
This is analogous to the context gating \cite{miech2017learnable} in video analysis which scales the input $\alpha^t$ (since $cgate(\h^{t-1}) \in (0,1)$) based on some context, which in our setup is the previous time points.

In the context of $J_\vv$ computation in Eq.~\eqref{eq:jacobian}, we observe that this `auxiliary' variable $\h^{t-1}$ could safely be used to transform $\uu_1$ without altering the triangular nature of the Jacobian matrix by (1) learning an element-wise operation $cgate(\h^{t-1})$ on $\uu_1$ for diagonal ${\partial \vv_1}/{\partial \uu_1}$ which (2) is not a function of $\uu_2$ so ${\partial \vv_1}/{\partial \uu_2} = 0$.
Thus, we now have
\begin{equation}
\small
    J_\vv = \begin{vmatrix} \frac{\partial \vv_1}{\partial \uu_1} & \frac{\partial \vv_1}{\partial \uu_2} \\
    \frac{\partial \vv_2}{\partial \uu_1} & \frac{\partial \vv_2}{\partial \uu_2}
    \end{vmatrix}
    = \begin{vmatrix}
    diag(cgate(\h^t)) & 0 \\
    \frac{\partial \vv_2}{\partial \uu_1} & diag(\exp s(\uu_1))
    \end{vmatrix}
\end{equation}
where $J_\vv = [\prod_j cgate(\h^t)_j]*[\exp(\sum_i (s(\uu_1))_i)]$.

As seen in Fig.~\ref{fig:network}, we place $f_{\textrm{TCG}}$ to the originally non-transformed variable of each \textit{layer} of a block (i.e., $\uu_2$ in the bottom layer and $\vv_1$ in the top layer).
We specifically chose a gating mechanism for conservative adjustments so that the original information is preserved to a large degree through simple but learnable `weighting'.
The full forward and inverse steps involving $f_{\textrm{TCG}}$ can easily be formulated by following Eq.~\eqref{eq:rec-coupling} and Eq.~\eqref{eq:inverse-rec-coupling} while respecting the order of operations seen in Fig.~\ref{fig:network}. See appendix for details.

\subsection{How do we use \ourmodel{}?}
Before we demonstrate the results using \ourmodel{} on the experiments, let us briefly describe how we can use \ourmodel{} for conditional sequence generation and density estimation.
In essence, \ourmodel{} aims to model an invertible mapping $[\{\y^t\},\{\z^t\}] = f(\{\x^t\})$ between sequential/longitudinal measures $\{\x^t\}$ and their corresponding observations $\{\y^t\}$ with $\{\z^t\}$ encoding the latent information across $t=1,\dots,T$ time points.
Once we train $f(\cdot)$, we can perform the following exemplary tasks:

\textbf{(1) Conditional sequence generation:} Given a series of observations of interest $\{\y^t\}$, we can sample $\{\z^t\}$ (each independently from a standard normal distribution) to generate $\{\x^t\} = f^{-1}([\{\y^t\},\{\z^t\}])$. The advantage comes from how $\{\y^t\}$ can be flexibly constructed (either seen or unseen from the data) such as an arbitrary disease progression over time.
Then, we randomly generate corresponding measurements $\{\x^t\}$ to observe the corresponding longitudinal measurements for both quantitative and qualitative downstream analyses. Since the model is recurrent, the sequence length can be extended beyond the training data to model the future trajectory.

\textbf{(2) Sequential density estimation:} Conversely, given  $\{\x^t\}$, we can predict $\{\y^t\}$, and more importantly, estimate the density $p_\X(\{\x^t\})$ at each $t$.
When $\{\x^t\}$ is generated from $\{\y^t\}$, the estimated density can indicate the `integrity' of the generated sample (i.e., low $p_\X$ implies that sequence is perhaps less common with respect to $\{\y^t\}$).

In the following section, 
we demonstrate several experiments that \ourmodel{} is able to  precisely perform these tasks described above. 

\section{Experiments}
To validate our framework in both qualitative and quantitative manners, 
we demonstrate two sets of experiments (one from a image sequence dataset and the other from a neuroimaging study) that are carried by successfully generating conditional sequences of data 
and estimating sequential densities on two separate datasets. 

\begin{figure}[!t]
	\vspace{-15pt}
	\centering
	\raisebox{2\height}{Train: \hspace{4.9pt}} \includegraphics[width=0.85\columnwidth]{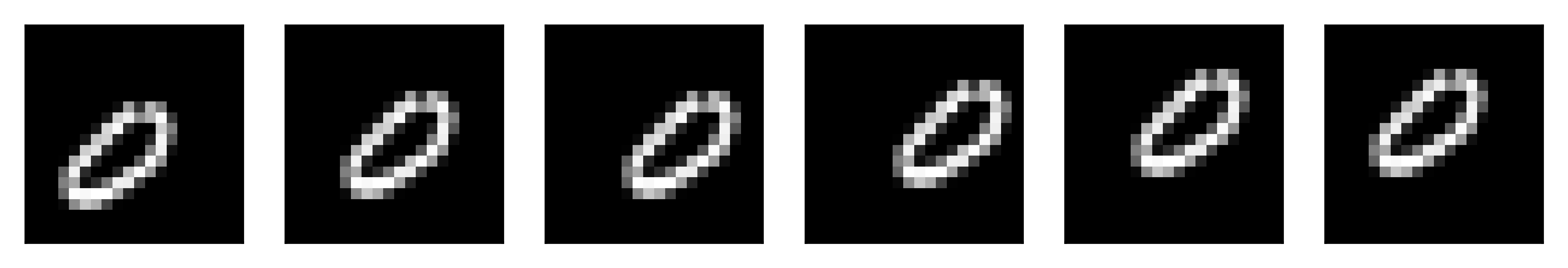} \\
	\raisebox{2\height}{cINN: \hspace{2.3pt}} \includegraphics[width=0.85\columnwidth]{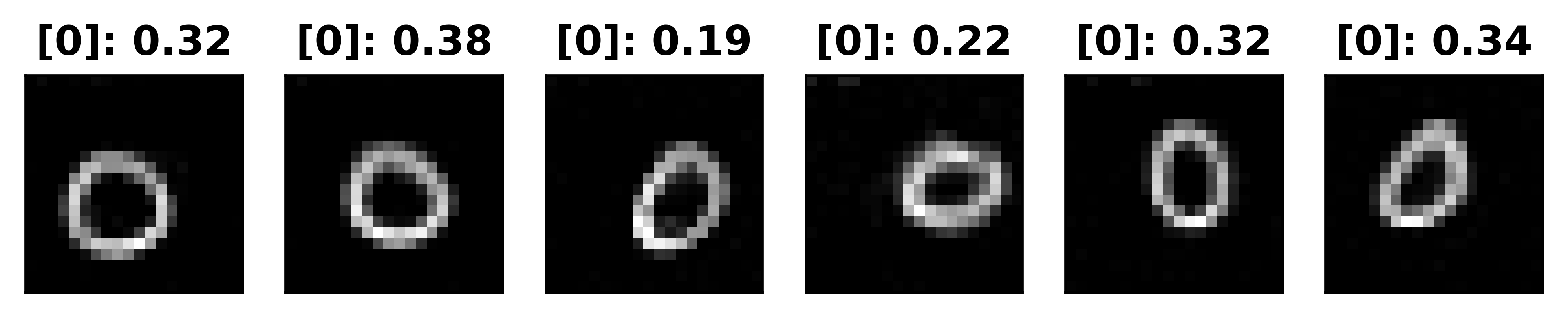} \\
	\raisebox{2\height}{Ours: \hspace{6.1pt}} \includegraphics[width=0.85\columnwidth]{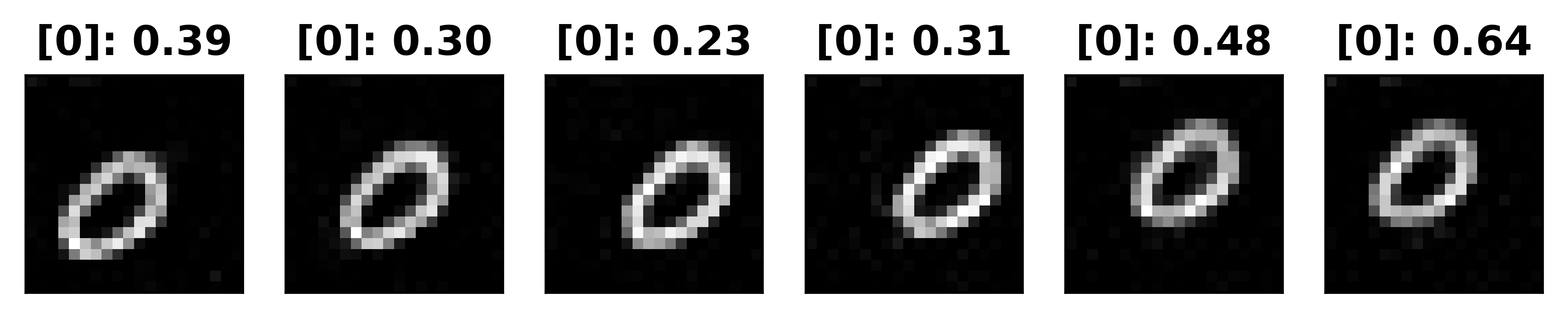}
	\vspace{-20pt}
	\caption{Top: A training sequence example. Middle: A generated 0-to-0 sequence example using cINN. Bottom: A generated 0-to-0 sequence example using our model. ([digit label]: estimated density)}
	\label{fig:mnist1}
\end{figure}

\subsection{Conditional Moving MNIST Generation}
We first test our model on a controlled moving MNIST dataset \cite{srivastava2015unsupervised} that consists of sequences of images showing a hand-written digit from 0 to 9 starting moving from left to right and oscillating its trajectories once hitting the boundary. 
This experiment qualitatively shows that a sequence of images generated with specific conditions (i.e., image labels) look smooth between sequences. 
Here, we specifically construct $\sim$13K controlled sequences of length $T=6$ where each frame of a sequence is an image of size 20 by 20  (vectorized as $\x^t \in \R^{400}$) and a one-hot vector $\y^t \in \R^2$ labeling the digit at $t$ indicating two different digit values (i.e., $[1,0]$ and $[0,1]$ respectively).

Our model consists of three coupling layers as shown in Fig.~\ref{fig:network}. 
An input is split into two halves $\uu_1$ and $\uu_2$, 
and each layer performs two successive transformations on each half once. 
In each transformation, 
one half of the input are passed through temporal contextual gating as defined in Eq.~\eqref{eq:cgate} 
and the other half through transformation blocks that contain one GRU cell and three layers of residual fully connected networks with ReLU non-linearity. 

Models were trained on 6 time points, but since the model is recurrent, further time points can be generated.
The labels for each sequence of images were ``identical'', i.e., $\y_t = [1,0]$ for $t=1,\dots,6$ for training. 
An example of training image sequences of 0 is given at the top panel of Fig.~\ref{fig:mnist1}. 

The middle and bottom panels in Fig.~\ref{fig:mnist1} show a primitive comparison of 
images sequences generated using cINN and our method. 
While the sequences from cINN are independent from each other, i.e., sampled independently under the same condition, 
the generated images using \textit{\ourmodel{}} show smooth and natural ``progress'' across the sequence. 


\begin{figure}[t]
	\centering
	\raisebox{2\height}{Ours: \hspace{2pt}} \includegraphics[width=0.87\columnwidth]{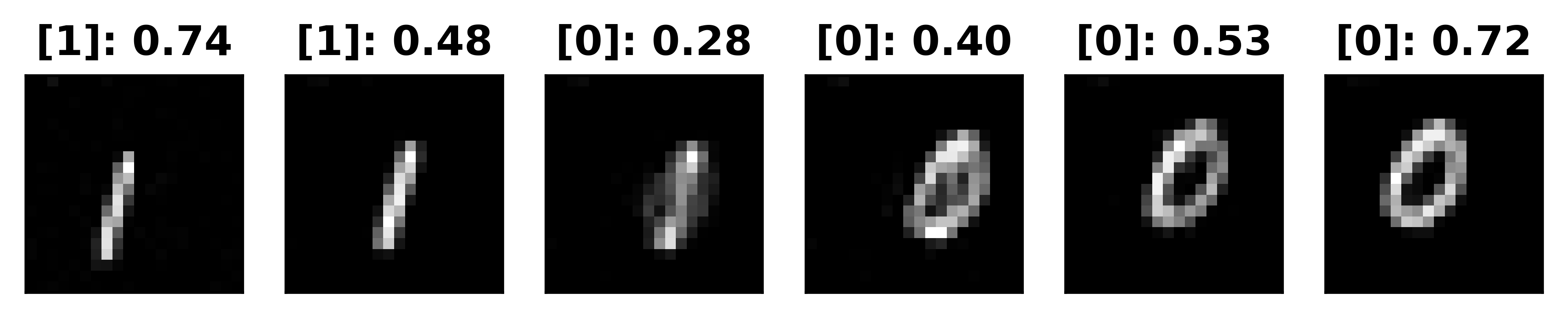}
	\raisebox{2\height}{cINN:} \hspace{1pt} \includegraphics[width=0.87\columnwidth]{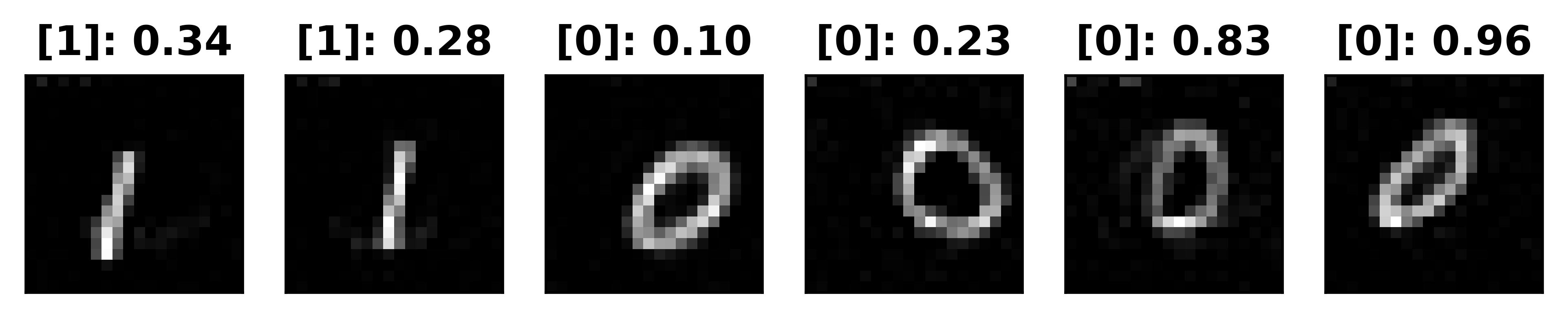}\hspace{-1pt}
	\vspace{-18pt}
	\caption{Examples of generated sequences using flipped observations. The number at the top of images show likelihood of the image belonging to the class. 
		Top: 1-to-0 using our model. Bottom: 1-to-0 using cINN. 
		Images from our model shows smooth transition while image using cINN show drastic change in the sequence. }
	\label{fig:good_2}
	\vspace{-15pt}
\end{figure} 

\begin{figure}[!t]
	\centering
	\raisebox{2\height}{5-to-3:} \includegraphics[width=0.85\columnwidth]{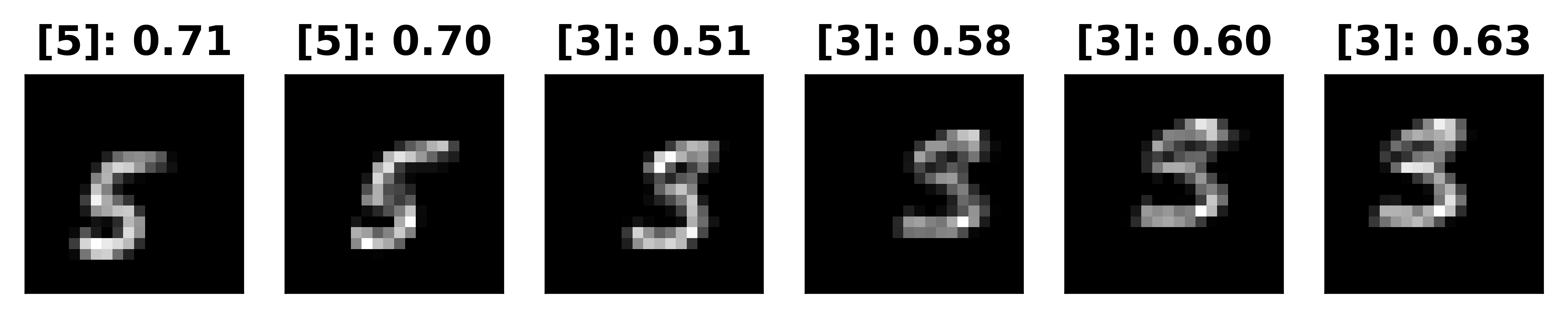}
	\raisebox{2\height}{3-to-5:} \includegraphics[width=0.85\columnwidth]{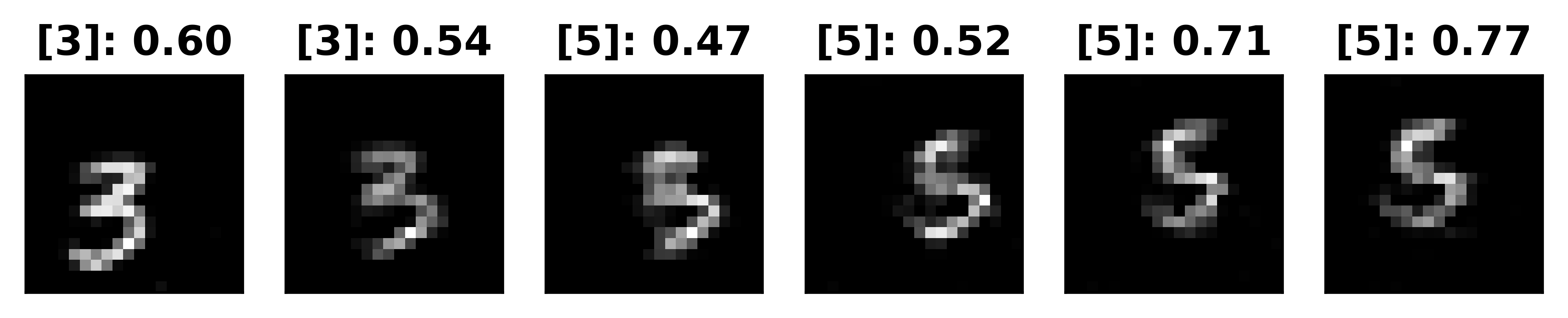}
	\raisebox{2\height}{9-to-5:} \includegraphics[width=0.85\columnwidth]{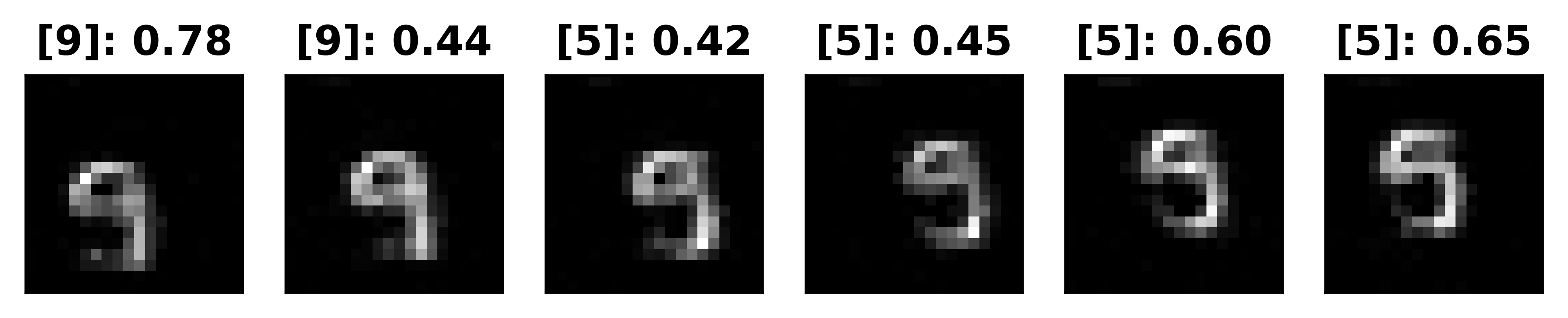}
	\raisebox{2\height}{5-to-9:} \hspace{1.5pt}\includegraphics[width=0.85\columnwidth]{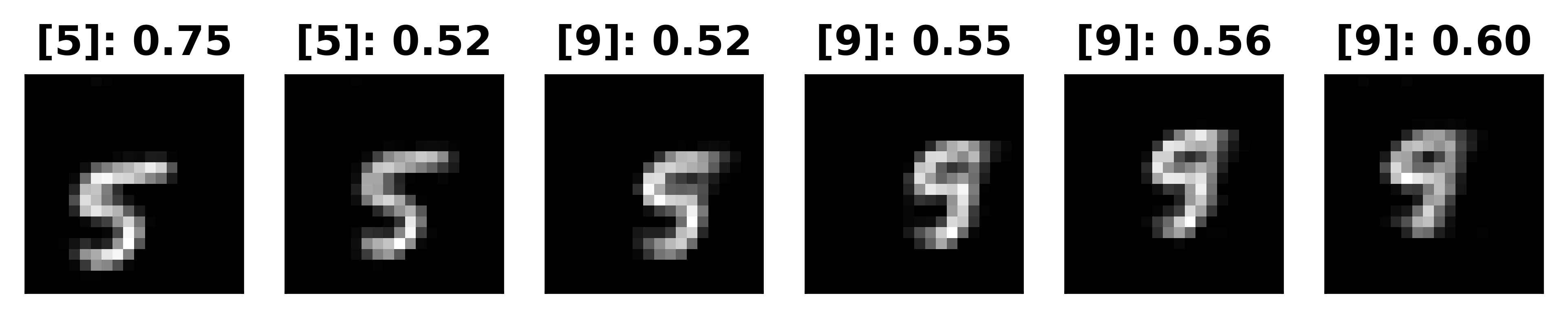}
	\vspace{-5pt}
	\caption{Examples of generated sequences with condition changes using our model. 
		Top: 3-to-5,  
		Bottom: 5-to-9.}
	\vspace{-5pt}
	\label{fig:good_3}
\end{figure}

In the next step, we changed the condition (i.e., image label) during a sequence generation (e.g., 0-to-1 or 5-to-3)  
to visually check if the changes between the sequences look natural. 
Note that the model was originally trained to learn the sequence of images with the same number.
Despite this setting was unseen during the training phase, we expected that the image sequence generated using our model 
exhibit ``smooth'' progress of changes. 
One demonstrative result is shown in Fig.~\ref{fig:good_2} where we compare the generated sequence of images with 
condition (i.e., label) changes from 1 to 0. 
Images using our model at the bottom of Fig.~\ref{fig:good_2} show gradual transition of the number in the middle of the sequence, 
while the images generated using cINN does not show such behavior.
Furthermore, our model quantifies its output confidence in the form of density (i.e., likelihood) shown at the top of each generated images in Fig.~\ref{fig:good_2}.  
Not only our model adjust generation based on inputs, 
it also outputs relatively low density at first when encountering the change as such patterns were not observed during the training, 
i.e., the likelihood decreases when condition changes and then increase as the sequences goes. 
This means that our model not only shows the conditional generation ability, 
it also estimates outputs' relative density given the training data seen. 
Different from other generative models, 
it allows conditional generation on sequential data while maintaining exact and efficient density estimation. 
More figures to visualize qualitative results are shown in Fig.~\ref{fig:good_3} that shows natural transition from one number to the other.

\begin{table*}[t]
	\centering
	\setlength{\tabcolsep}{7pt} 
	\renewcommand{\arraystretch}{1.2} 
	\begin{tabular}{l|ccccc}	
		&  Diagnosis            & ADAS13       & MMSE         & RAVLT-I        & CDR-SB        \\ \hline \hline
		Control     & CN$\rightarrow$CN$\rightarrow$CN  & 10$\rightarrow$10$\rightarrow$10 & 30$\rightarrow$30$\rightarrow$30 & 70$\rightarrow$70$\rightarrow$70 & 0$\rightarrow$0$\rightarrow$0  \\
		Progression           & CN$\rightarrow$MCI$\rightarrow$AD & 10$\rightarrow$20$\rightarrow$30 & 30$\rightarrow$26$\rightarrow$22 & 70$\rightarrow$50$\rightarrow$30 & 0$\rightarrow$5$\rightarrow$10 \\ \hline
		cINN    ($N_1$=100 / $N_2$= 100)      & 4             & 2            & 0            & 0            & 0          \\
		Ours    ($N_1$=100 / $N_2$= 100)     & 11            & 12           & \textbf{2}   & \textbf{2}   & 7          \\
		Ours + TCG ($N_1$=100 / $N_2$= 100)  & \textbf{12}   & \textbf{14}  & \textbf{2}   & \textbf{2}   & \textbf{9} \\ \hline \hline
		Control &            CN$\rightarrow$CN$\rightarrow$CN  & 10$\rightarrow$10$\rightarrow$10 & 30$\rightarrow$30$\rightarrow$30 & 70$\rightarrow$70$\rightarrow$70 & 0$\rightarrow$0$\rightarrow$0  \\
		Early-progression            & CN$\rightarrow$MCI$\rightarrow$MCI & 10$\rightarrow$13$\rightarrow$16 & 30$\rightarrow$28$\rightarrow$26 & 70$\rightarrow$60$\rightarrow$50 & 0$\rightarrow$2$\rightarrow$4  \\ \hline
		cINN  ($N_1$=150 / $N_2$= 150)        &  0             & 2            & 0            & 0            & 0          \\
		Ours  ($N_1$=150 / $N_2$= 150)         & 2             & 4            & \textbf{4}   & \textbf{1}   & 0          \\
		Ours + TCG ($N_1$=150 / $N_2$= 150)    & \textbf{4}    & \textbf{5}   & \textbf{4}   & \textbf{1}   & \textbf{1}\\	
		\hline
	\end{tabular}
	\vspace{-5pt}
	\caption{Number of ROIs identified by statistical group analysis using the generated measures with respect to various covariates associated with AD.  
		Each column represents sequences of disease progression represented by diagnosis or test scores. 
		\textit{\ourmodel{}} considers the progression sequences while cINN generates cross-sectional data in different conditions.
		In all cases, using \textit{\ourmodel{}} with TCG yields the most number of statistically significant ROIs.  
	} 
	\label{table:neuroimaging}
\end{table*}
\subsection{Longitudinal Neuroimaging Analysis}

The goal of this neuroimaging experiment is to validate if our conditionally generated samples actually exhibit statistically robust and clinically sound characteristics 
when trained with a longitudinal AD brain imaging dataset. 
After training, we generate {\em sufficient} number of longitudinal brain imaging measures (i.e., $\{\x^t\}$) conditioned on various covariates (i.e., labels $\{\y^t\}$) associated with AD progression.
These data in sequence should show ``progress'' of pathology consistent with the covariates.  
Then, with the generated sequences conditioned on two chosen groups (e.g., healthy group vs. disease progressing group), 
we perform a statistical analysis to detect disease related features of the data measurements. 
In the end, we expect that regions of interests (ROIs) identified by the statistical group analysis are consistent with other AD literature 
with statistically stronger signal (i.e., lower $p$-value) than results using the original training data.

\textbf{Dataset.} The Alzheimer's Disease Neuroimaging Initiative (ADNI) database (\url{adni.loni.usc.edu}) is one of the largest and still growing neuroimaging databases.
Originated from ADNI, we use a longitudinal neuroimaging dataset called The Alzheimer's Disease Prediction of Longitudinal Evolution (TADPOLE) \cite{marinescu2018tadpole}. 
We used data from $N$$=$$276$ participants with $T=3$ time points.  

\begin{figure}[!t]
	\centering
	\includegraphics[height=70pt]{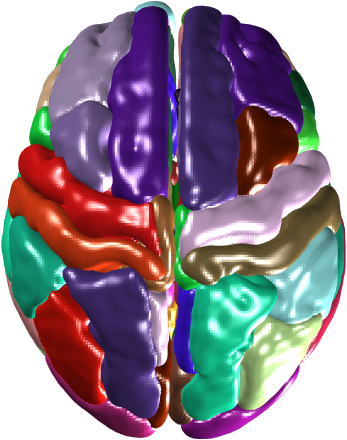}
	\hspace{10pt}
	\includegraphics[height=70pt]{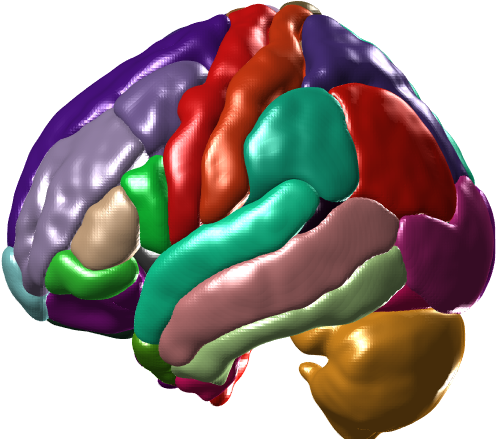}
	\hspace{10pt}
	\includegraphics[height=70pt]{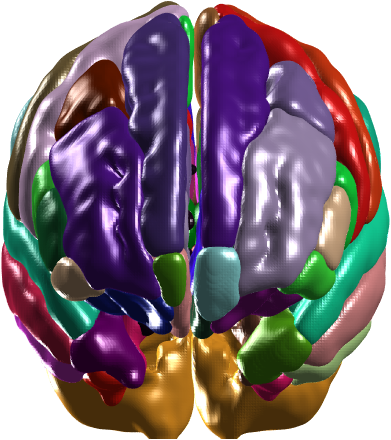}
	\caption{Left to right: Top, front and side view of Desikan brain atlas. Different ROIs have different colors.}
	\label{fig:desikan-atlas}
	\vspace{-10pt}
\end{figure}

\textbf{Input.} For the longitudinal brain imaging sequence $\{\x^t\}$, we chose Florbetapir (AV45) Positron Emission Tomography (PET) scan measuring the level of amyloid-beta deposited in  brain which has been a known type of pathology associated with Alzheimer's disease \cite{wong2010vivo,joshi2012performance}.
The AV45 images were registered to a common brain template (MNI152) to derive the gray matter regions of interests (82 Desikan atlas ROIs \cite{desikan2006automated}, see Fig.~\ref{fig:desikan-atlas}).
Thus, each ROI entry of $\x^t \in \R^{82}$ holds an average Standard Uptake Value Ratio (SUVR) measure of AV45 where high AV45 implies more amyloid pathology.

\textbf{Condition.} For the corresponding labels $\{\y^t\}$ for longitudinal conditions, we chose five covariates with known associations to AD progression (parentheses for normal to impaired range): (1) \textit{Diagnosis}: Normal/Control (CN), Mild Cognitive Impairment (MCI) and Alzheimer's Disease (AD). (2) \textit{ADAS13}: Alzheimer's Disease Assessment Scale (0 to 85). (3) \textit{MMSE:} Mini Mental State Exam (0 to 30). (4) \textit{RAVLT-I:} Rey Auditory Verbal Learning Test - Immediate (normal (0 to 75). (5) \textit{CDR:} Clinical Dementia Rating (0 to 18).
These assessments impose disease \textit{progression} during the generation step. See supplement and \cite{marinescu2018tadpole} for details.

\textbf{Analysis.}
We performed a statistical group analysis on each condition $\{\y^t\}$ independently with the following pipeline: \textit{(1) Training:} First, we trained our model (same subnetworks as the Moving MNIST experiment model) using 
SUVR $\x^t \in \R^{82}$ for 82 ROIs with $\{\y^t\}$ as the `labels'. \textit{(2) Conditional longitudinal sample generation:} Then, we generate longitudinal samples $\{\ \hat{\x}^t \}$ conditioned on two distinct longitudinal conditions: Control (Healthy across all sequences) versus Progress (condition gets worse).

For each condition (e.g., Diagnosis), we generate $N_1$ samples of Control (e.g., $\{\hat{\x_1}^t \}$ conditioned on $\{\y_1^t \} =$ CN$\rightarrow$CN$\rightarrow$CN) and $N_2$ samples of Progress ($\{\hat{\x_2}^t \}$ conditioned on $\{\y_2^t \} =$ CN$\rightarrow$MCI$\rightarrow$AD).
Then, we perform a two sample t-test at $t=3$ between each entry of $\{\hat{\x_1}^3$ and $\{\hat{\x_2}^3$ and derive its multiple testing corrected $p$-value.

\begin{figure*}[!t]
	
	\centering
	\begin{minipage}{230pt}\centering
		\includegraphics[height=70pt]{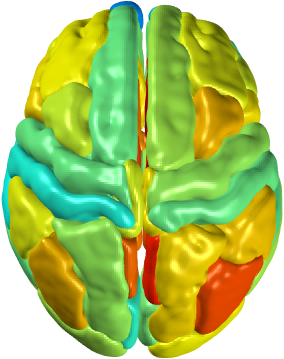} \hspace{3pt}
		\includegraphics[height=70pt]{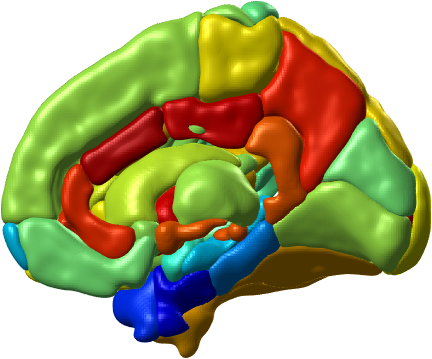} \hspace{3pt}
		\includegraphics[height=70pt]{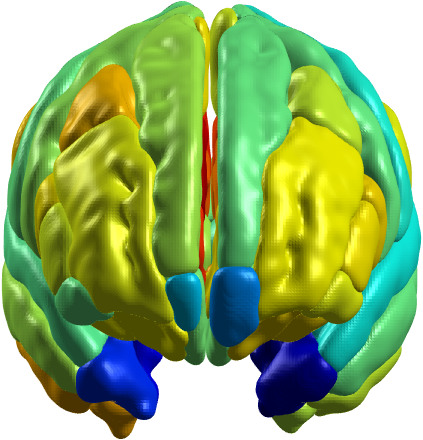} \hspace{3pt} \\ \vspace{2pt}
		\includegraphics[height=70pt]{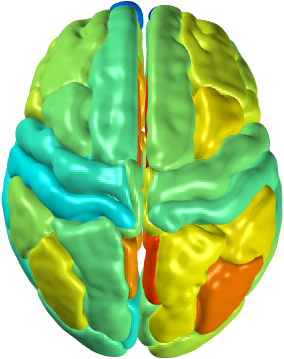} \hspace{3pt}
		\includegraphics[height=70pt]{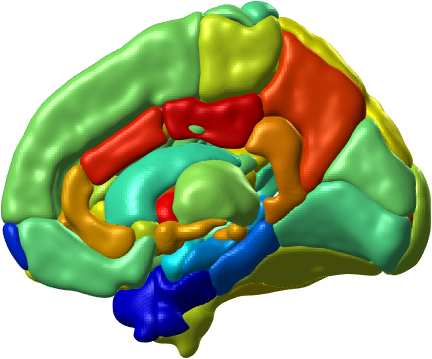} \hspace{3pt}
		\includegraphics[height=70pt]{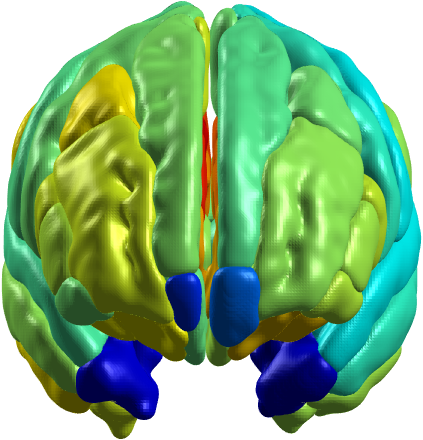} \hspace{3pt} \\ \vspace{2pt}
		\includegraphics[height=70pt]{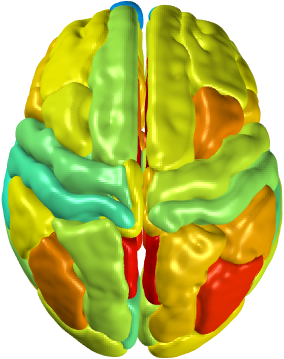} \hspace{3pt}
		\includegraphics[height=70pt]{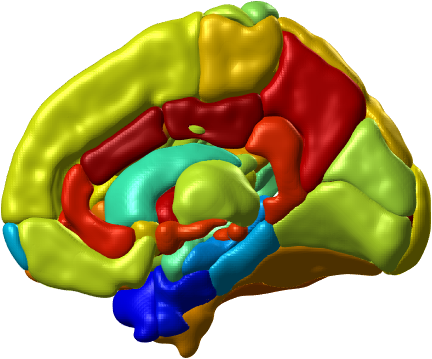} \hspace{3pt}
		\includegraphics[height=70pt]{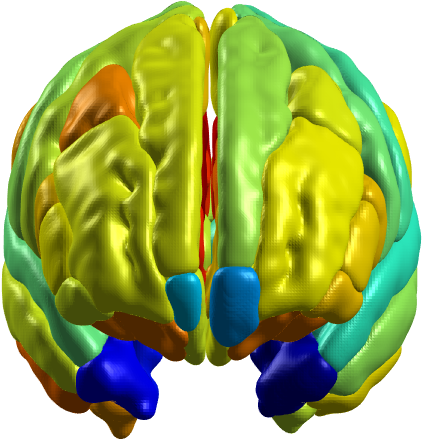} \hspace{3pt} 
	\end{minipage}\hspace{10pt}
	\begin{minipage}{250pt}\centering
		\includegraphics[height=70pt]{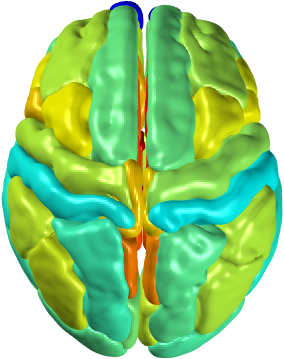} \hspace{3pt}
		\includegraphics[height=70pt]{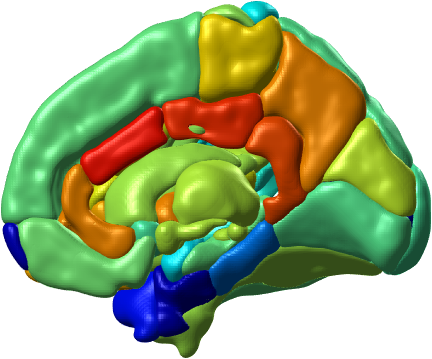} \hspace{3pt}
		\includegraphics[height=70pt]{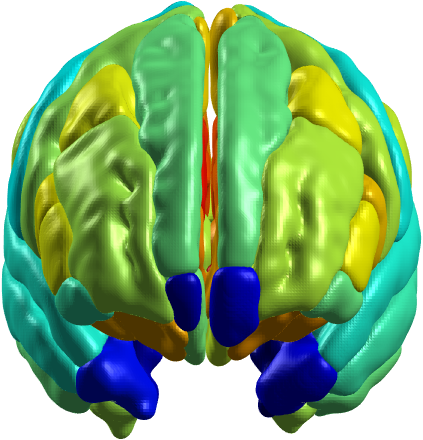} \hspace{3pt} 
		\includegraphics[height=70pt]{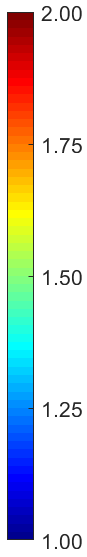} \\ \vspace{2pt}
		\includegraphics[height=70pt]{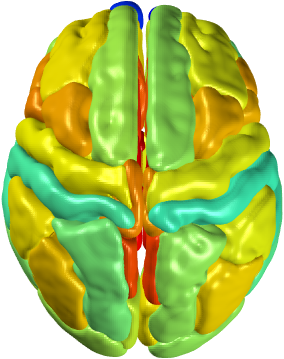} \hspace{3pt}
		\includegraphics[height=70pt]{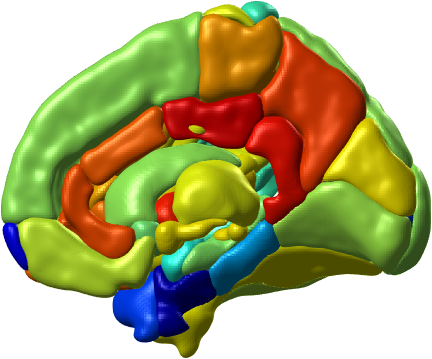} \hspace{3pt}
		\includegraphics[height=70pt]{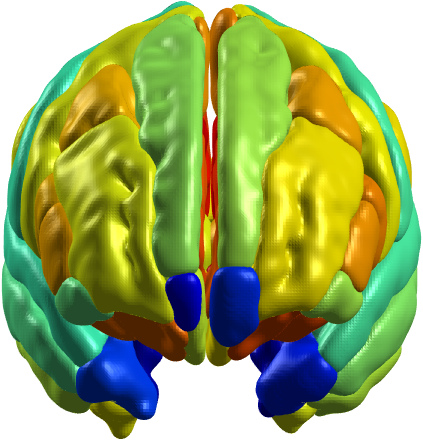} \hspace{3pt} 
		\includegraphics[height=70pt]{figs/colorbar.png} \\ \vspace{2pt}
		\includegraphics[height=70pt]{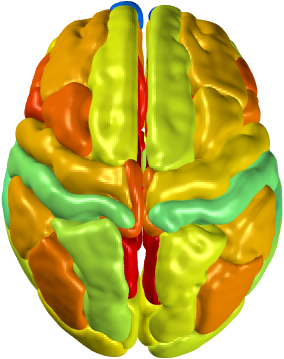} \hspace{3pt}
		\includegraphics[height=70pt]{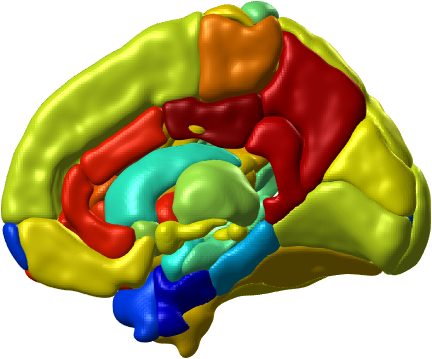} \hspace{3pt}
		\includegraphics[height=70pt]{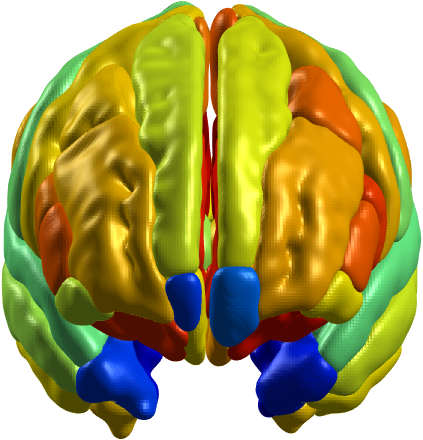} \hspace{3pt} 
		\includegraphics[height=70pt]{figs/colorbar.png} 
	\end{minipage}
	
	\caption{Training data vs. generated samples comparison for CN (top)$\rightarrow$MCI (middle)$\rightarrow$AD (bottom). \textbf{Left three columns:} The average of the samples CN$\rightarrow$MCI$\rightarrow$AD trajectory in the dataset. \textbf{Right three columns:} The average of the 100 generated sequences conditioned on CN$\rightarrow$MCI$\rightarrow$AD. Red and blue indicate high and low AV45 respectively, from top to bottom, ROIs are expected to turn more red (i.e., disease progression). We observe that the generated samples (right 3 columns) show magnitudes and sequential patterns similar to those of the real samples from the training data.}
	\label{fig:trajectory}
\end{figure*}

\begin{figure}[!t]
	\centering
	\includegraphics[height=62pt]{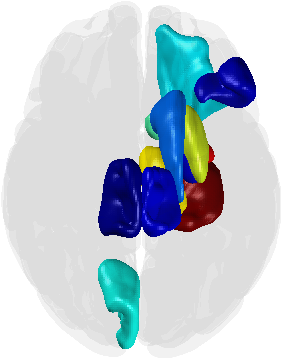} \hspace{3pt}
	\includegraphics[height=62pt]{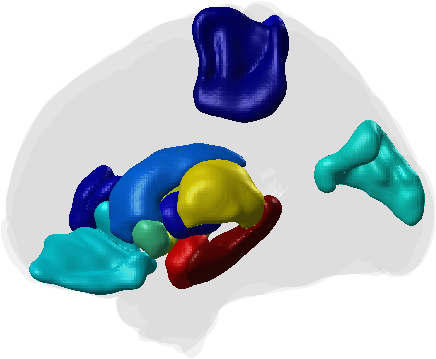} \hspace{3pt}
	\includegraphics[height=62pt]{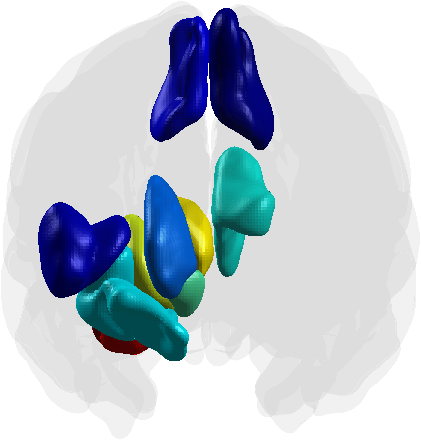} \hspace{3pt}
	\includegraphics[height=62pt]{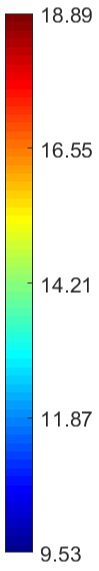} \hspace{3pt}
	\caption{12 ROIs found to be significantly different between two Diagnosis groups ( CN$\rightarrow$CN$\rightarrow$CN vs. CN$\rightarrow$MCI$\rightarrow$AD) at $t=3$ using our model under `Diagnosis' in Table~\ref{table:neuroimaging}. The colors denote the -log $p$-value. Several AD-related ROIs such as hippocampus, putamen, caudate and amygdala are included.}
	\label{fig:pvalue}
	\vspace{-15pt}
\end{figure}

\textbf{Result.}
\textit{Control vs. Progression (Top row block of Table~\ref{table:neuroimaging}):} We set longitudinal conditions for each covariate based on its associated to healthy progression (e.g., low ADAS13 throughout) and disease progression (e.g., high ADAS13 related to eventual AD onset).
We generated $N_1 = 100$ and $N_2=100$ samples for each group respectively.
Then, we performed the above statistical group difference analysis under 4 setups: (1) Raw training data, (2) cINN \cite{ardizzone2018analyzing}, (3) Our model, and (4) Our model + TCG.
With out sampling, the sample size of the desirable longitudinal conditions was extremely small for all setups, so no statistical significances were found.
With cINN, we occasionally found few significant ROIs, but the non-sequential samples with only $t=3$ could not generate realistic samples.
With \ourmodel{} we consistently found significant ROIs
Further, \ourmodel{} with the temporal context gating (TCG) detected the most number of ROIs which we visualize in Fig.~\ref{fig:pvalue}.

\textit{Control vs. Early-progression (Bottom row block of Table~\ref{table:neuroimaging}):}
We setup a more challenging chase where we generate samples which resemble subjects that show slower progression of the disease (i.e., lower rate of covariate change over time).
Such case is especially important in AD when early detection leads to effective preventions.
By following the same statistical analyses set as the Control vs. Progression setup, we first sampled $N_1=100$ and $N_2=100$ samples, but no significant ROIs were found in any of the models.
To improve the sensitivity, we generated $N_1=150$ and $N_2=150$ samples in all models and found several significant ROIs only with \ourmodel{} (bottom rows of Table~\ref{table:neuroimaging}).

\textit{Generation assessment:} In Fig.~\ref{fig:trajectory}, we see the generated samples (right 3 columns) through time (top to bottom) in three views of the ROIs and compare them to the real training samples (left 3 columns).
We see that the generated samples have similar AV45 loads through the ROIs, and more importantly, the progression pattern across time (i.e., ROIs turning more red indicating amyloid accumulation) follows that of the real sequence as well.
Also, throughout the analyses setups, the significant ROIs often involved AD-related regions acknowledged in the neuroscience field, this implies that the generated longitudinal sequences consistently follow the underlying distribution of the real data which we may not have been able to make use of otherwise.

\section{Conclusion}
\vspace{-5pt}
Motivated by various experimental setups in recent computer vision and neuroimaging studies 
that require large-scale longitudinal/temporal sequence of data analysis, 
we study the problem of 
generative models using neural networks that accounts for 
progressive behavior of longitudinal data sequences. 
By developing a novel architecture of an invertible neural network 
that incorporates recurrent subnetworks and temporal context gating 
to pass information within a sequence generation, 
we enable a neural network to ``learn'' the conditional distribution of training data in a latent space 
and generate a sequence of samples that demonstrate progressive behavior according to the given conditions 
such as different levels of diagnosis labels (e.g., healthy to MCI to AD).
We demonstrate exhaustive experimental results with various experimental settings using two datasets 
that validate such longitudinal progress in sequential image generation and AD pathology. 
Especially in neuroimaging applications which often suffer from small sample sizes, 
our results show promising evidence 
that our model can provide sufficient number of generated samples 
to obtain statistically robust results. The code will be publicly available on \url{https://github.com/shwang54}.

{\small
\bibliographystyle{splncs}
\bibliography{cvpr19-sjh}
}
\begin{appendices}
\newpage
\appendixpage
In this appendix, we provide the following additional details:

\begin{enumerate}
	\item \textbf{Training details:} We provide additional details on how we trained our model along with the invertible neural network (loss functions). Additional Jacobian determinant explanation is also provided.
	\item \textbf{Alzheimer's disease dataset:} We also provide more information on the Alzheimer's disease dataset (ADNI and TADPOLE) and the covariates (disease related conditions such as ADAS13) used in the experiments.
	\item \textbf{Experiment setup details:} For improved reproducibility, we provide the exact setup of the experiments such as hyper-parameters and network structures.
	\item \textbf{Neuroimaging experiment result details:} Details of the neuroimaging experiments are provided including the ROI information of the detected significant ROIs.
\end{enumerate}

\section{Training Details}
\subsection{Loss functions}
There are three loss functions for the bidirectional training \cite{ardizzone2018analyzing} described in the main text. Without loss of generality, let us consider a single time point which can simply be extended to multiple time points by computing these losses to each of the entire time points.
\begin{enumerate}
	\item $\mathcal{L}_\Z ( p(\y,\z), p(\y)p(\z) )$: This is a loss in the forward mapping. Specifically, given a input $\x$, we first forward map it to $[\hat{\y},\z] = f(\x)$ which corresponds to $p(\hat{\y},\z)$ as our network maps both $\y$ and $\z$ with a single network. Our goal is to minimize the distance between this distribution resulting from our network ($p(\hat{\y},\z)$) to the ideal joint distribution $p(\y)p(\z)$. But since we may not exactly know $p(\y)$ and $p(\z)$, a kernel-based moment matching measure called Maximum Mean Discrepancy (MMD) \cite{dziugaite2015training} is used which only uses the samples without explicitly requiring $p(\y)$ and $p(\z)$. Specifically, for each $\x$ and its corresponding forward map $[\hat{\y},\z]$, we also construct its ``counterpart'' sample $[\y_{gt},\z \sim \Z]$ which is simply the ground truth $\y_{gt}$ and a random sample $\z$ from a standard normal $\Z$. In other words, we construct a set of samples representing the joint distribution $p(\y)p(\z)$ by empirically setting $\y_{gt}$ and a sample $\z$ from the true prior $\Z$ which we have been assuming. Thus, the loss is fully expressed in practice as follows:
	\begin{align}
	\mathcal{L}_\Z ( p(\y,\z), p(\y)p(\z) ) &= MMD([\hat{\y},\z]_{i=1}^N \\
	&= f(\x_{i=1}^N),[ (\y_{gt})_{i=1}^N, \z_{i=1}^N \sim \Z] )
	\end{align}
	for $N$ samples in each mini-batch. The kernel used in the MMD is an inverse multiquadratic kernel 
	\[
	k(\x,\x') = \frac{\alpha}{\alpha  + ||\x - \x'||_2^2}
	\]
	where we used $\alpha = \{ 0.2, 0.5, 0.8, 1.0, 1.2 \}$ for multiple scales of $\alpha$ \cite{ardizzone2018analyzing,tolstikhin2017wasserstein}.
	\item $\mathcal{L}_\Y(\y_{gt}, \hat{\y})$: This is another loss in the forward mapping. Similar to typical supervised loss, it penalizes the difference between the true $\y_{gt}$ and the predicted $\hat{\y}$. We used the mean squared error (MSE) function.
	\item $\mathcal{L}_\X(p(\x), p_\X)$: This is a loss in the inverse mapping (hence, the bidirectional training together with the above losses). Intuitively, this enforces the reconstructed $\x_{reconst}$ with known $\y$ ($p(\x)$) and random $\z$ to follow a likely $\x$ with the same $\y$ ($p_\X$). Instead of maximizing the log likelihood of $p(\x)$ directly, this is again achieved via MMD that for a given set of $\x_{i=1}^N$ (and their $\y_{i=1}^N$), we construct a set of samples with random $\z$ and the same set of $\y_{i=1}^N$ to perform the kernel-based distance measure. We use the same kernel function (and $\alpha$'s) as $\mathcal{L}_\Z$.	
\end{enumerate}
For all these losses, the ratios of the terms were all equal throughout the experiments

\subsection{Jacobian determinant and Temporal Context Gating}
Here, we provide further explanations to the determinant of Jacobian computation that we promised in the main text along with the adjustment for the Temporal Context Gating.
\subsubsection{Chain rule of determinants}
When our network (or any other involving coupling layers) involves multiple coupling layers, the Jacobian determinant of the entire network can be computed by computing the individual Jacobian determinants. Let us consider only one coupling layer for now. We know that the Jacobian determinant of each coupling layer can be computed as the product of the diagonal entries of the Jacobian since the Jacobian is a square triangular matrix of the following form (without temporal context gating):
\begin{equation}\label{eq:jacobian}
\small
J_\vv = \frac{\partial \vv}{\partial \uu} = \begin{vmatrix} \frac{\partial \vv_1}{\partial \uu_1} & \frac{\partial \vv_1}{\partial \uu_2} \\
\frac{\partial \vv_2}{\partial \uu_1} & \frac{\partial \vv_2}{\partial \uu_2}
\end{vmatrix}
= \begin{vmatrix}
I & 0 \\
\frac{\partial \vv_2}{\partial \uu_1} & diag(\exp s(\uu_1))
\end{vmatrix}
\end{equation}
thus $J_\vv = \exp(\sum_i (s(\uu_1))_i)$ and with the temporal context gating, it is
\begin{equation}\label{eg:jacobian-tcg}
\small
J_\vv = \begin{vmatrix} \frac{\partial \vv_1}{\partial \uu_1} & \frac{\partial \vv_1}{\partial \uu_2} \\
\frac{\partial \vv_2}{\partial \uu_1} & \frac{\partial \vv_2}{\partial \uu_2}
\end{vmatrix}
= \begin{vmatrix}
diag(cgate(\h^t)) & 0 \\
\frac{\partial \vv_2}{\partial \uu_1} & diag(\exp s(\uu_1))
\end{vmatrix}
\end{equation}
where $J_\vv = [\prod_j cgate(\h^t)_j]*[\exp(\sum_i (s(\uu_1))_i)]$.
Here, we note that the upper left block of $J_\vv$ in Eq.~\eqref{eg:jacobian-tcg} is diagonal since each output of $cgate(\h^t)$ is multiplied to each element of $\uu_1$ in an element-wise manner. This behavior is identical to how the lower right block $J_\vv$ in Eq.~\eqref{eg:jacobian-tcg} and Eq.~\eqref{eq:jacobian} is simply $diag(\exp s(\uu_1))$ due to the element-wise product transform applied to $\uu_1$.

The Jacobian determinants of the subsequent coupling layers can be computed consecutively using the output of the previous coupling layer as the input to the current coupling layers. In other words, for a series of composited formulations $f = f_1 \circ f_2 \circ \cdots \circ f_N$ where each $f_i$ is a coupling layer operation, then $\det(f)$ is
\begin{align}
\det(f) &= \det(f_1 \circ f_2 \circ \cdots \circ f_N) \\
&= \det(f_1) \det(f_2) \cdots \det(f_N)\\
&= \prod_{i=1}^N \det(f_i).
\end{align}
This allows an easy computation of the full Jacobian determinant regardless of the number of coupling layer operations.

\subsection{Basic training details}
We used NVIDIA 1080ti GPU to train all the models. ADAM optimizer with $\alpha=0.9$ and $\beta=0.999$ and the initial learning rate of $0.0005$ was used.

\section{Alzheimer's Disease Dataset}
We provide additional details regarding the Alzheimer's disease dataset we used. These are largely based on Alzheimer's Disease Neuroimaging Initiative (ADNI) databse (\url{adni.loni.usc.edu}) and The Alzheimer's Disease Prediction of Longitudinal Evolution (TADPOLE) \cite{marinescu2018tadpole}.

\subsection{Dataset (ADNI and TADPOLE)}
\subsubsection{Pre-processing}
The images were pre-processed with standard ADNI pipelines. For PET images, including AV45, were co-registered, averaged across the dynamic range, standardized with respect to the orientation and voxel size (see \url{adni.loni.usc.edu/methods/pet-analysis/pre-processing} for full details). For AV45, standardized uptake value ratio (SUVR) measures for Desikan ROIs \cite{desikan2006automated} after segmented and parcellated with Freesurfer (version 5.3.0) were measured (see \cite{jagust2010alzheimer} for full details).

\subsubsection{TADPOLE Datasets}
TADPOLE consists of multiple datasets, each serving different purposes with respect to the challenge itself. For our experiments, we used D1 and D2: (i) D1 is the standard training set consisting of individuals with at least two separate visits across three phases of the ADNI study (ADNI1, ADNI GO and ADNI2) and (ii) D2 is the longitudinal prediction set which have the rollovers (i.e., subjects from D1 with further visits) for the purpose of forecasting tasks. In our setup, we simply treated the subjects in D1 and D2 without distinctions to obtain the most number of subjects with (i) 3 time points with (ii) AV45 measures for all 3 time points and (iii) covariates of interests at each time point.

\subsubsection{Covariates (Conditions) Description}
Here, we describe the covariates (conditions) we used in the experiments to characterize the disease progression from varying perspectives. Full documents are available on \url{http://adni.loni.usc.edu/methods/documents} (e.g., ADNI Procedures Manual).
\begin{enumerate}
	\item \textbf{Diagnosis}: This is the diagnosis by each visit code. There are largely three categories: Control (CN, healthy), Mild Cognitive Impairment (MCI, combining early and late MCI) and Alzheimer's Disease (AD, diseased). In TADPOLE, it is provided with the baseline diagnosis along with the per-visit diagnosis change (DX) to indicate the change from previous diagnosis to the current diagnosis. We explicitly used the diagnosis change code to assign the current status (e.g., if DX is CN$\rightarrow$MCI, we assign MCI to the current diagnosis). Note that although it might be logical to think that the disease progresses monotonically (e.g., MCI$\rightarrow$AD but not AD$\rightarrow$MCI), the diagnosis is only with respect to the corresponding visit, so such revert cases are present in the dataset, capturing the progression especially more difficult.
	\item \textbf{ADAS13}: Alzheimer's Disease Assessment Scale - Cognitive (ADAS-COG) is a set of 13 assessments for learning and memory, language production, language comprehension, constructional parxis, ideational parxis, and orientation. The values (in the current dataset) range from 0 to 85 where higher number indicates worse cognitive functions (e.g., more number of errors made). While ADAS13 is not a physical measure of AD (compared to, for instance, amyloid), this measure the level of cognitive deficit which is a well-known symptom of dementia. Note that dementia is not specific to AD; thus more AD-specific pathologies such as amyloid and tau are observed with respect to these cognitive measures.
	\item \textbf{MMSE}: Mini Mental State Exam (MMSE) consists of a series of tasks testing the cognitive impairment. The score ranges from 0 to 30 where higher score indicates better cognitive function (opposite from ADAS13). MMSE has been used for reliable diagnosis of AD.
	\item \textbf{RAVLT-I}: Rey Auditory Verbal Learning Test - Immediate (RAVLT-I) tests learning ability which has been shown to be associated with AD \cite{moradi2017rey}. The score ranges from 0 to 75 where lower score indicates lower cognition.
	\item \textbf{CDR-SB}: Clinical Dementia Rating - Sum of Boxes (CDR-SB) measures scores from six different categories such as memory, orientation, judgment and problem solving and community affairs. The score ranges from 0 to 18 (from out samples) where higher score indicates higher dementia rating. This especially prioritizes memory functional over other categories.
\end{enumerate}

\section{Experiment Details}
Here, we provide specifics of the experiments such as the feature dimensions, feature splitting and network dimensions.

\begin{figure*}[!t]
	\centering
	\includegraphics[height=130pt]{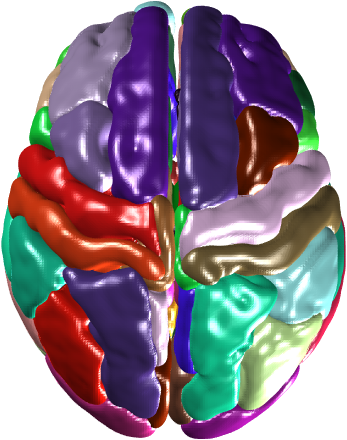}
	\includegraphics[height=130pt]{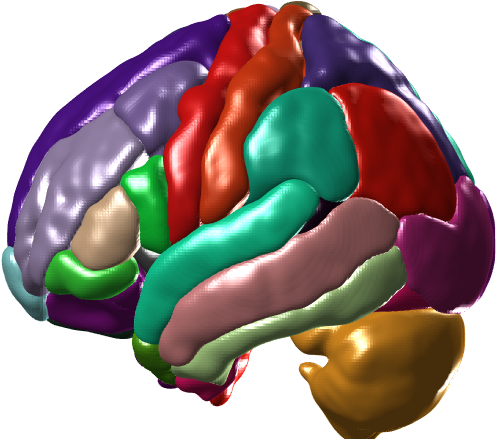}
	\includegraphics[height=130pt]{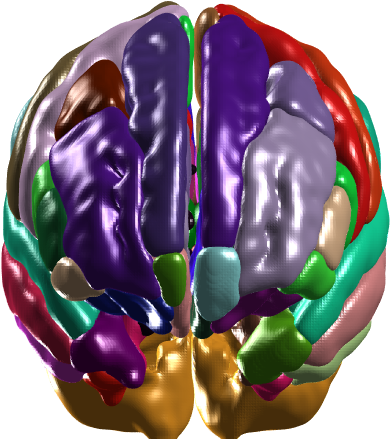}\\
	\includegraphics[height=130pt]{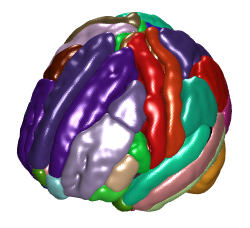}
	\includegraphics[height=130pt]{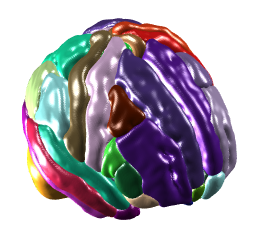}
	\caption{Desikan atlas ROIs. Top row (from left): top, side and front view. Bottom row: diagonal views.}
	\label{fig:desikan}
\end{figure*}
\subsection{Moving MNIST}
\begin{enumerate}
	\item Input/output dimensions: An input $\x^t$ at each time point $t$ was $\R^{400}$ (image of size 20 by 20 was vectorized). The condition $\y^t$ was $\R^2$ (onehot vector for 2 digit case) and the latent variable $\z^t$ was $\R^8$ which was chosen by us. For a smaller sized $\z^t$, the densities could not be accurately captured. On the other hand, for a larger sized $\z^t$, the latent information could potentially ``memorize'' the input to output mapping which is also undesirable. Both input and output were zero-padded to 512 dimensions.
	\item Input partitioning: We split $\x$ (i.e., $\uu$ without zero padding) into $[\x_1,\x_2]$ ($\x_1$ into $\uu_1$ and $\x_2$ into $\uu_2$) in a ``checkerboard'' pattern. In other words,  given an image $\x$, the first half $\x_1$ consists of the pixels that are not directly adjacent to each other (i.e., black squares in a checkerboard) and the remaining pixels (which are also not directly adjacent to each other like the white squares in a checkerboard) are assigned to $\x_2$. This splitting scheme preserves the overall geometric structure of the image as much as possible in a simplistic manner as did Real-NVP \cite{dinh2016density}.
	\item Network setup: We used 3 coupling blocks (2 coupling layers in each block) where the input/output dimensions (i.e., $\uu$ and $\vv$ dimensions) are 512 (thus, zero padding of length 112 is needed for $\x^t$ and 502 is needed for $[\y^t,\z^t]$). Each subnetwork $q$ then has the input/output dimensions of 256 where it first starts with a GRU (256 input, 256 output, 256 hidden) followed by 3 residual layers (fully connected layers with ReLU non-linearity, all of 256 input and 256 output).
\end{enumerate}

\subsection{Neuroimaging experiments}
\begin{enumerate}
	\item Input/output dimensions: An input $\x^t$ at each time point $t$ is $\R^{82}$ for 82 brain regions. The condition $\y^t$ is $\R^3$ for diagnosis (onehot-vector over three possible diagnosis categories) and $\R^1$ for other continuous covariates. The latent variable $\z^t$ was $\R^4$. Both input and output were zero-padded to 150 dimensions.
	\item Input partitioning: We simply split the input $\x^t$ by assigning the left hemisphere features to $\x_1^t$ and the right hemisphere features to $\x_2^t$ (there are exactly 41 left and 41 right hemisphere features). This was the most logical setup since the brain regions are symmetrical.
	\item Network setup: The setup is exactly the same as the Moving MNIST setup except for the input/output zero-padded dimensions of 150.
\end{enumerate}

\section{Neuroimaging Experiment Result Details}

\subsection{Desikan ROIs}
Fig.~\ref{fig:desikan} is the enlarged version of the Fig.~6 in the main text (bottom two with diagonal views) showing the Desikan ROIs. The colors distinguish different ROIs (not reflecting any measures).
Note that the orange regions (cerebellums) are not shown in the AV45 figures since AV45 are not measured in those regions (standard procedure for treating cerebellums in other modalities as well).

\subsection{Significant ROIs}
Below, we show the list of the significant ROIs presented in Table 1 of the main text. These are the predefined names of the Desikan atlas \cite{desikan2006automated}. Some naming conventions are as follows: CTX: cortex, LH: left hemisphere, RH: right hemisphere. Significance threshold of $\alpha = 0.01$ with the Bonferroni multiple testing correction ($\alpha / 82$) was used for all the setups to control for family-wise error rate.

\subsubsection{AV45 vs. Diagnosis (CN$\rightarrow$CN$\rightarrow$CN vs. CN$\rightarrow$MCI$\rightarrow$AD)}
\begin{enumerate}
	\setlength\itemsep{0.01em}
	\item CTX$\_$LH$\_$LINGUAL
	\item CTX$\_$LH$\_$PARAHIPPOCAMPAL
	\item CTX$\_$LH$\_$PERICALCARINE
	\item LEFT$\_$AMYGDALA
	\item LEFT$\_$CAUDATE
	\item LEFT$\_$HIPPOCAMPUS
	\item LEFT$\_$PALLIDUM
	\item LEFT$\_$PUTAMEN
	\item LEFT$\_$THALAMUS$\_$PROPER
	\item LEFT$\_$VENTRALDC
	\item CTX$\_$RH$\_$ENTORHINAL
	\item CTX$\_$RH$\_$PARAHIPPOCAMPAL
\end{enumerate}

\subsubsection{AV45 vs. ADAS13 (10$\rightarrow$10$\rightarrow$10 vs. 10$\rightarrow$20$\rightarrow$30)}
\begin{enumerate}
	\setlength\itemsep{0.01em}
	\item CTX$\_$LH$\_$FUSIFORM
	\item CTX$\_$LH$\_$LATERALOCCIPITAL
	\item CTX$\_$LH$\_$PERICALCARINE
	\item LEFT$\_$HIPPOCAMPUS
	\item LEFT$\_$PALLIDUM
	\item LEFT$\_$THALAMUS$\_$PROPER
	\item LEFT$\_$VENTRALDC
	\item CTX$\_$RH$\_$CUNEUS
	\item CTX$\_$RH$\_$FUSIFORM
	\item CTX$\_$RH$\_$LATERALOCCIPITAL
	\item CTX$\_$RH$\_$LATERALORBITOFRONTAL
	\item CTX$\_$RH$\_$LINGUAL
	\item CTX$\_$RH$\_$PERICALCARINE
	\item CTX$\_$RH$\_$SUPERIORTEMPORAL
\end{enumerate}

\subsubsection{AV45 vs. MMSE (30$\rightarrow$30$\rightarrow$30 vs. 30$\rightarrow$26$\rightarrow$22)}
\begin{enumerate}
	\setlength\itemsep{0.01em}
	\item CTX$\_$LH$\_$ISTHMUSCINGULATE
	\item CTX$\_$RH$\_$FRONTALPOLE
\end{enumerate}

\subsubsection{AV45 vs. RAVLT-I (70$\rightarrow$70$\rightarrow$70 vs. 70$\rightarrow$50$\rightarrow$30)}
\begin{enumerate}
	\setlength\itemsep{0.01em}
	\item CTX$\_$LH$\_$ISTHMUSCINGULATE
	\item CTX$\_$LH$\_$PARACENTRAL
\end{enumerate}

\subsubsection{AV45 vs. CDR-SB (0$\rightarrow$0$\rightarrow$0 vs. 0$\rightarrow$5$\rightarrow$10)}
\begin{enumerate}
	\setlength\itemsep{0.01em}
	\item CTX$\_$LH$\_$PERICALCARINE
	\item LEFT$\_$HIPPOCAMPUS
	\item LEFT$\_$PALLIDUM
	\item LEFT$\_$THALAMUS$\_$PROPER
	\item LEFT$\_$VENTRALDC
	\item CTX$\_$RH$\_$CUNEUS
	\item CTX$\_$RH$\_$LATERALOCCIPITAL
	\item CTX$\_$RH$\_$LINGUAL
	\item CTX$\_$RH$\_$PERICALCARINE
\end{enumerate}


\end{appendices}

\end{document}